\documentclass{article}
\usepackage{fontspec}

\usepackage{iftex}
\ifLuaTeX
  \usepackage{luatex85}
\fi

\usepackage{fontspec}
\PassOptionsToPackage{numbers,sort&compress}{natbib}
\AtBeginDocument{\let\cite\citep}

\usepackage[eandd, final]{neurips_2026}

\usepackage[utf8]{inputenc} %
\usepackage[T1]{fontenc}    %
\usepackage{hyperref}       %
\usepackage{url}            %
\usepackage{booktabs}       %
\usepackage{amsfonts}       %
\usepackage{nicefrac}       %
\usepackage{microtype}      %
\usepackage[table]{xcolor}  %
\usepackage{booktabs}
\usepackage{graphicx}
\usepackage{enumitem}
\usepackage{wrapfig}
\usepackage{fancyvrb}
\usepackage{fvextra}
\usepackage[skins,breakable]{tcolorbox}
\usepackage{makecell}
\usepackage{array}

\usepackage{enumitem}
\ifLuaTeX
  \newfontfamily\promptpreviewfont{fonts/FreeSerif.ttf}[NFSSFamily=nbpromptpreview,Renderer=HarfBuzz]
\else
  \newfontfamily\promptpreviewfont{fonts/FreeSerif.ttf}[NFSSFamily=nbpromptpreview]
\fi

\DefineVerbatimEnvironment{PromptVerbatim}{Verbatim}{
  breaklines=true,
  breakanywhere=true,
  breaksymbolleft={},
  breaksymbolright={},
  fontsize=\scriptsize,
  formatcom=\promptpreviewfont
}

\newtcolorbox{promptbox}[1]{
  breakable,
  enhanced,
  title={#1},
  colback=gray!2,
  colframe=blue!40!black,
  colbacktitle=blue!15,
  coltitle=blue!45!black,
  fonttitle=\bfseries,
  boxrule=0.45pt,
  arc=1mm,
  left=1.2mm,
  right=1.2mm,
  top=1mm,
  bottom=1mm,
  before skip=0.75em,
  after skip=0.75em
}

\providecommand{\NBResultsPath}{NeurIPS_2026}
\graphicspath{{\NBResultsPath/}}
\providecommand{\nbsymbolimg}[2][0.28cm]{\ensuremath{\vcenter{\hbox{\includegraphics[height=#1]{#2}}}}}

\providecommand{\nbAgency}{\nbsymbolimg{icons/agency.png}}
\providecommand{\nbFilter}{\nbsymbolimg{icons/filter.png}}
\providecommand{\nbEcho}{\nbsymbolimg{icons/echo.png}}
\providecommand{\nbLanguage}{\nbsymbolimg{icons/language.png}}

\title{\textsc{The Narrative Gap}: Can LLMs Help Us\\ Navigate Diverse Narratives Across Languages?}

\author{Nikhil Sharma$^{\dagger}$, Kelly Marchisio$^{\ddagger}$, Kenton Murray$^{\diamondsuit}$, Ziang Xiao$^{\dagger}$ \\
$^{\dagger}$Johns Hopkins University  ~~~~~ $^{\ddagger}$ Cohere ~~~~~ $^{\diamondsuit}$George Mason University \\
\texttt{nsharm27@jhu.edu, kelly@cohere.com, kenton@jhu.edu ,ziang.xiao@jhu.edu}}

\begin{document}

\maketitle

\begin{abstract}
Large language models (LLMs) increasingly mediate how people seek and consume information, including accounts that differ across languages and regions, especially on contested topics. Supporting diverse information seeking requires models to help users navigate competing narratives across languages and information streams. We define six capabilities necessary for diverse multilingual information seeking: three foundational capabilities for narrative inclusion, narrative exclusion, and missing narrative identification; and three higher-level capabilities for presenting diverse narratives, preserving narrative coverage under confirmatory queries, and maintaining information parity across languages. To guide the development of those information seeking capabilities, we introduce \textsc{NarrativeBench}, a benchmark comprising 1,333 news articles in 23 publication languages across 23 contested topics and 460 Reddit-grounded queries. Evaluating 14 LLMs across 24 query languages, we find that current models lack the capabilities needed to reliably navigate diverse narratives, limiting users' control over the perspectives they receive and leaving substantial gaps in exposure to competing accounts even when diversity is explicitly requested. Information access also differs across query framings and languages, risking echo chambers, linguistic filter bubbles, and information disparities that weaken the common ground needed for democratic discourse. \textsc{NarrativeBench} provides a foundation for developing information systems that give users across languages and regions agency to navigate diverse narratives.

\par\smallskip
{\small\centering
\setlength{\tabcolsep}{3pt}
\begin{tabular}{@{}cll@{}}
\raisebox{-0.25em}{\includegraphics[height=1.35em]{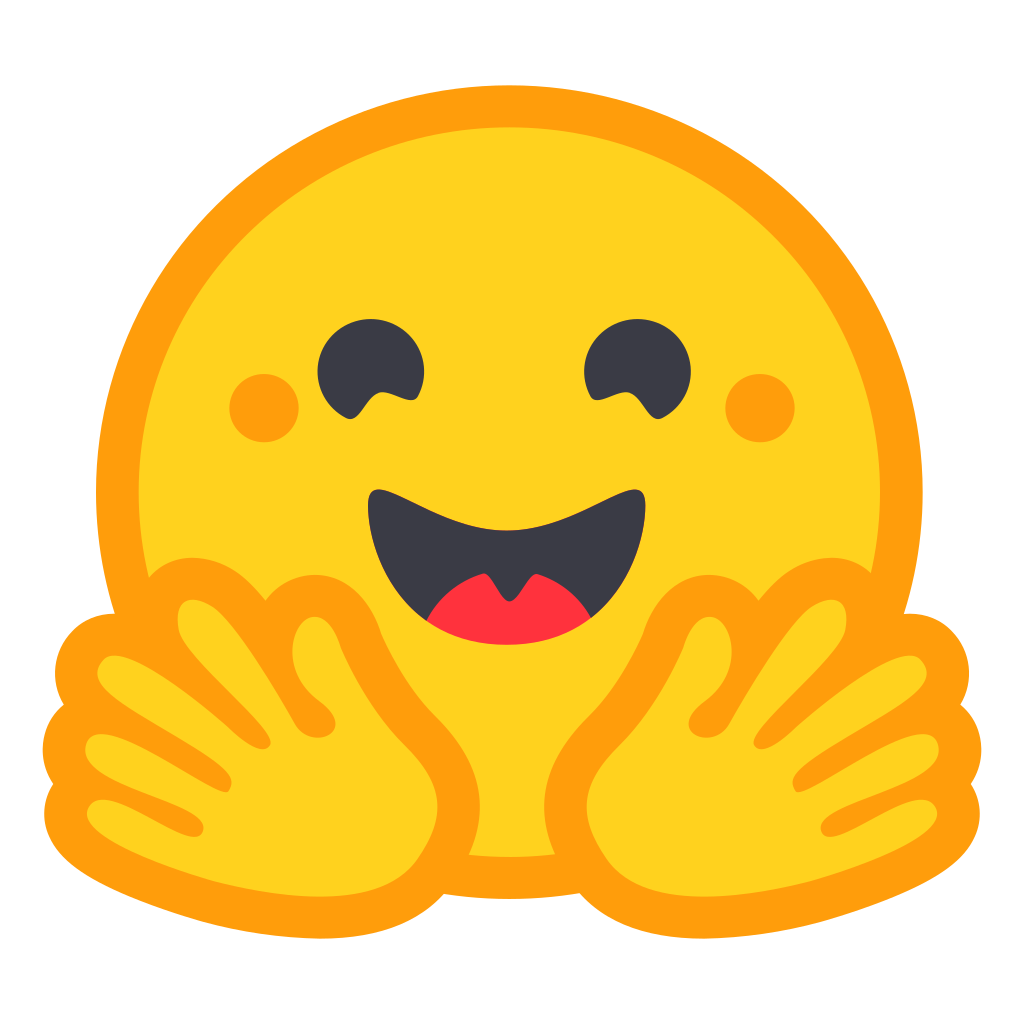}} & \textbf{Dataset:} & \url{https://huggingface.co/datasets/jhu-clsp/narrative-bench} \\[2pt]
\raisebox{-0.25em}{\includegraphics[height=1.35em]{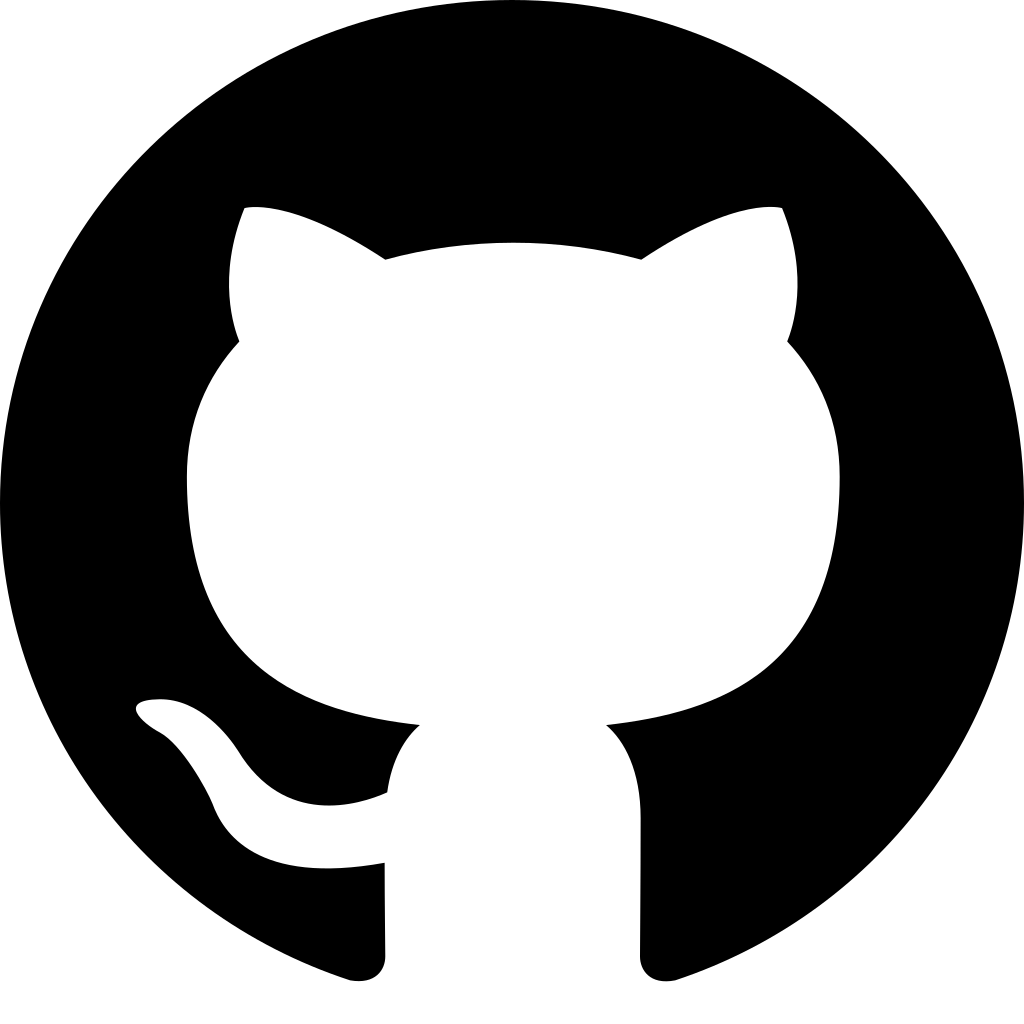}} & \textbf{Code:} & \url{https://github.com/isle-dev/Narrative-Bench-NeurIPS-2026}
\end{tabular}\par
}

\end{abstract}

\section{Introduction} 
We, informavores \citep{miller1984informavores}, seek and consume information continually to find what we need to know, decide, or act on. 
Despite information today being more accessible, navigation has grown harder. New information is produced at a scale and speed no information seeker can track, and what exists is fragmented across geographic and linguistic silos, each with its own vocabulary, claims, framings, and emphases.
Throughout history, we have built information systems, from libraries to search engines and social media feeds, to help us navigate the complex information world. 
However, these systems often reproduce, and sometimes amplify, the very problems they were meant to solve. 
By organizing content around a user's query language, region, and prior behavior, they create filter bubbles \citep{10.5555/2029079, 10.5555/2361740, csis-gaza-through-whose-lens-filter-bubbble} and echo chambers \citep{doi:10.1073/pnas.2023301118,alatawi2021survey}, increasing polarization and weakening the common ground necessary for democratic stability \citep{common-knowledge-attack-on-democracy}. 
For an information seeker, overcoming these barriers is costly: it requires knowing which foreign perspectives to search for, translating conflicting sources, and identifying missing vocabulary and information silos.

Large Language Models (LLMs) hold promise as a new kind of information system. 
In principle, their multilingual, conversational, and reasoning capabilities should let a user ask a question in any language from any region and receive an answer that synthesizes across the perspectives available, bridging cultural divides and improving democratic deliberation \citep{ai-common-ground}. 
But are LLMs actually helping us navigate this complex information world? 
A growing body of research auditing linguistic, political, and cultural biases in LLMs suggests they may instead reproduce and exacerbate the barriers they are meant to overcome. 
Those models create generative echo chambers by reinforcing user beliefs \citep{generative-echo-chamber, 10.1145/3786304.3787879,nehring2024large,sharma2023towards,doi:10.1126/science.aec8352,ibrahim2026warm}, prefer information in the query language over more relevant foreign sources \citep{faux-polyglot,https://arxiv.org/abs/2509.13930}, and struggle to reason when evidence contains explicit conflicts \citep{NEURIPS2024_baf4b960, Shi_Jin_Shen_Dong_Wu_Xiong_2024, https://aclanthology.org/2024.emnlp-main.486/,faux-polyglot}. Whether LLMs can serve as information systems that break, rather than reinforce, the traditional barriers of information seeking remains an open question.

In this paper, we identify three foundational information seeking capabilities that give users control over which narratives an LLM presents: requesting a particular narrative, excluding a narrative, and identifying perspectives missing from an answer. Building on the foundational capabilities, we evaluate three higher-level information seeking capabilities: presenting diverse narratives, preserving narrative coverage under confirmatory queries, and maintaining information parity across languages.
We evaluate the capabilities across contested topics, including geopolitical conflicts where the stakes of navigation failure are high.
Recent global conflicts underscore the urgency: Israel–Palestine, India–Pakistan, and Russia–Ukraine are not only fought through military and diplomatic action, but also through parallel information environments in which different regional and linguistic communities encounter different accounts of the same events \citep{csis-gaza-through-whose-lens-filter-bubbble, pak-india-info-war, gao2023wmgic}. 
Information flows in such geopolitical conflicts consist of culturally situated accounts with differing causal and moral interpretations of the same event \citep{Entman_1993, doi:10.1177/1464884916680372}, making information seeking challenging.
Information seekers in such environments need to navigate and parse diverging information streams across languages, making translation alone insufficient.
We formulate this challenge as the problem of navigating diverse multilingual narratives.
In this paper, narratives are defined as culturally situated accounts of events that interpret the actions and intentions of those involved \citep{bruner1991narrative}, and operationalized as accounts of a topic expressed in a particular language by sources anchored in a geographic region.
However, we acknowledge that narratives vary even within the same geographic region and language. Our operationalization simplifies this complexity to provide an auditable unit for measuring how information systems allow users to navigate diverse information streams.
We argue that an information system's ability to navigate narratives as operationalized in this paper is a necessary but not sufficient condition for enabling diverse multilingual information seeking.
Future work should extend this operationalization to capture narrative differences within regions and languages.

To evaluate the foundational and higher-level information seeking capabilities, we introduce \textsc{NarrativeBench}, an ecologically grounded dataset and benchmark comprising 1,333 news articles in 23 publication languages across 23 contested topics and 460 Reddit-grounded queries. The evaluation covers 4,800 cases across 24 query languages (Appendix~\ref{appendix:current-evaluation-protocol}).

We find that current multilingual LLMs (1) struggle to distinguish and selectively present different perspectives, (2) fail to consistently present diverse views across languages, (3) vary in narrative coverage between neutral and confirmatory queries, and (4) perform unevenly across languages. Such capability gaps decrease exposure to diverse narratives and hinder effective information seeking, potentially exacerbating linguistic filter bubbles and echo chambers, reducing common ground across regions, and amplifying existing conflicts and divisions in societies.

Overall, this paper makes the following contributions:
\begin{itemize}[leftmargin=*,topsep=4pt,itemsep=2pt,parsep=0pt,partopsep=0pt]
\raggedright
    \item We formulate diverse multilingual information seeking as a problem of navigating narratives and identify foundational and higher-level capabilities LLMs need to help users navigate diverse perspectives across languages and regions.
    \item We introduce \textbf{NarrativeBench}, an ecologically grounded dataset and benchmark to evaluate the information seeking capabilities of multilingual LLMs.
    \item We provide an empirical analysis of state-of-the-art LLMs, identifying capability gaps that hinder diverse information seeking across languages and regions.
\end{itemize}

NarrativeBench provides a capability-grounded foundation for building information systems that promote equitable information access, reduce polarization, and enhance democratic discourse.
\section{Related Work}

\subsection{Diverse Information Seeking and Perspective Exposure}
Information seeking depends on the information available and the perspectives users encounter, select, and trust. Research on selective exposure and partisan news consumption shows that some audiences concentrate their attention on politically aligned sources \citep{https://reutersinstitute.politics.ox.ac.uk/echo-chambers-filter-bubbles-and-polarisation-literature-review,doi:10.1126/sciadv.abn0083}; strong political beliefs and the perceived strength of an issue can further increase susceptibility to confirmation bias during news consumption \citep{https://dl.acm.org/doi/10.1145/3706598.3713873}. Efforts to broaden perspective exposure include interfaces that help users explore news and opinions across political positions \citep{10.1145/3637340}, but explicitly labeling the stance of news articles can intensify users' selection of viewpoints aligned with their beliefs \citep{10.1145/3274324}. System choices about information exposure can also influence political attitudes: reducing the prominence of posts expressing antidemocratic attitudes and partisan animosity in news feeds can reduce affective polarization \citep{https://www.science.org/doi/10.1126/science.adu5584}.
Experiments with LLM-mediated search show that users exhibit greater confirmation bias in their queries than with conventional search \citep{generative-echo-chamber}. Biased model responses can further influence users' judgments by reinforcing existing beliefs or shifting attitudes toward the model's position \citep{doi:10.1126/science.aec8352,10.1145/3544548.3581196}. Supporting diverse information seeking therefore requires LLMs to surface relevant missing perspectives even when users ask confirmatory questions and do not explicitly request alternative views.\looseness=-1

\subsection{LLM-Mediated Information Access and Pluralism}
LLMs are increasingly used for information seeking and decision making \citep{chatterji2025people}. LLM-mediated systems shape access to perspectives by selecting sources, synthesizing claims, and deciding which information to include in their responses. Research on AI search and news consumption shows that citations can concentrate on a small set of outlets \citep{https://arxiv.org/abs/2507.05301} and identifies concerns about accuracy, transparency, and shared understanding \citep{10.1145/3772318.3791484,https://dl.acm.org/doi/10.1145/3706598.3713804}. Existing information-seeking benchmarks evaluate the accuracy and usefulness of retrieved evidence, the completeness of information collection, and reasoning and clarification in expert consultations \citep{xi2025infodeepseekbenchmarkingagenticinformation,wong2026widesearch,kim2026seekergym,zhu2026gisabenchmarkgeneralinformationseeking,dongre2026mirage}. Complementary work examines how models identify and distinguish perspectives, represent narrative frames, and accommodate diverse human preferences \citep{Chen_Khashabi_Yin_Callison-Burch_Roth_2019,Nie_Omoomi_Flek_Zhao_Welch_2025,https://arxiv.org/abs/2404.16019,https://aclanthology.org/2025.findings-acl.477}.

Evaluating perspective diversity in information seeking requires examining how query language and the available evidence shape the perspectives presented in model responses. \citet{https://aclanthology.org/2024.naacl-long.213/} show that LLMs answer territorial-dispute questions differently depending on query language, and knowledge-conflict work studies how models behave when context conflicts with parametric memory or when retrieved passages contradict one another \citep{https://aclanthology.org/2024.emnlp-main.486/,https://arxiv.org/abs/2406.13805,https://arxiv.org/abs/2404.10198,auKAUJZMO6}. The focus on language-dependent positions and conflicting facts leaves open how models represent competing narratives that emphasize different facts and interpretations of the same event across languages and regions.

\subsection{Multilingual LLMs and Information Capabilities}
Navigating diverse narratives requires information capabilities that enable access to perspectives across languages. Multilingual LLMs offer a foundation for cross-cultural information seeking, yet current evaluations find inconsistent answers to semantically equivalent questions and performance differences across languages, scripts, and resource levels \citep{https://aclanthology.org/2024.naacl-long.22/,https://aclanthology.org/2025.naacl-long.394/,https://arxiv.org/abs/2407.01358,https://aclanthology.org/2025.ijcnlp-long.185/,chang2025chikhapolargescalemultilingualbenchmark}. Culturally grounded benchmarks evaluate everyday knowledge \citep{https://arxiv.org/abs/2406.09948} and knowledge and reasoning through local examination questions \citep{https://arxiv.org/abs/2411.19799}. Multilingual question-answering evaluations use information-seeking questions collected directly in diverse languages to assess passage selection and answer extraction \citep{clark-etal-2020-tydi}.

Access to diverse narratives across languages also depends on which sources models select and use in their responses. In multilingual retrieval-augmented generation, where models answer questions using retrieved sources, \citet{faux-polyglot} find preferences for documents in the query language during retrieval and generation, and for higher-resource languages during generation when query-language evidence is unavailable. Related work identifies preferences for high-resource languages, English, and Latin scripts \citep{https://aclanthology.org/2025.findings-acl.295,https://arxiv.org/abs/2604.20199}; models can also favor a source's language over its relevance \citep{https://arxiv.org/abs/2509.13930}. Multilingual benchmarks evaluate answer generation using Wikipedia sources \citep{https://aclanthology.org/2025.naacl-long.14/}, while healthcare audits examine language disparities in factual correctness and response quality \citep{https://arxiv.org/abs/2410.13671v3,https://dl.acm.org/doi/10.1145/3589334.3645643}. Language preferences can therefore shape which narratives remain available to users even when the evidence is multilingual, motivating evaluation of the diversity represented in the resulting answers. \textsc{NarrativeBench} builds on research in multilingual information access and conflicting evidence by evaluating whether models can selectively present and recover narratives, broaden exposure under confirmatory queries, and maintain comparable information access across languages.
\section{\textsc{NarrativeBench}: The Dataset}
\label{sec:dataset}

\paragraph{Gathering the Narratives} We construct \textsc{NarrativeBench} by collecting news articles through localized searches about contested topics. Dataset curation followed \textbf{four stages}: (1) selecting contested topics across regions and languages, (2) identifying regions and stakeholders relevant to each topic, (3) developing region-specific search terms in the relevant languages, and (4) collecting relevant articles and their URLs from the search results. The corpus contains 1,333 news articles from 43 countries, regions, and regional organizations across 23 topics and 23 publication languages. Within each topic, we group articles by source perspective and publication language, yielding 180 narrative groups. Article texts have a median length of 4,654 characters (mean = 5,977, SD = 5,283, range = 610--46,252). We describe the article collection in Appendix~\ref{appendix:dataset-details}.

\paragraph{Dataset Diagnostics} To assess whether the collection captures distinct narratives, we compared multilingual embeddings of all 1,333 article texts in their publication languages, restricting comparisons to articles about the same topic. Articles within the same narrative group were more similar than articles from different groups (mean cosine similarity difference = 0.047, 95\% bootstrap CI [0.044, 0.051], permutation $p < 0.001$). After restricting comparisons to articles published in the same language, articles from the same source perspective remained more similar on average than articles from different perspectives (Appendix~\ref{appendix:corpus-diagnostics}).

\paragraph{Sourcing Queries for Evaluation} To ground evaluation in naturally occurring information seeking, we assembled 460 queries, comprising 283 questions gathered from Reddit and 177 context-specific adaptations. Reddit queries were gathered through topic-specific searches; adaptations addressed 12 topics with limited query coverage and were reviewed for fidelity to the source question, topic, and intended framing. Two independent annotators rated a sample of 100 English queries on how likely a person would be to ask the question to an AI assistant, giving mean ratings of 4.09 and 4.89 out of 5; neither annotator rated a query as unlikely or very unlikely.

To test whether models preserve access to diverse narratives when queries favor an existing belief, we include 230 neutral and 230 confirmatory queries. Neutral queries request information without endorsing a particular answer, while confirmatory queries seek support for an expressed or implied belief, interpretation, solution, or outcome. Comparing narrative coverage across query categories tests the robustness needed to limit reinforcement of echo chambers during information seeking~\citep{generative-echo-chamber,nickerson1998confirmation}.\looseness=-1

\paragraph{Query Translation} To compare information capabilities across query languages, we provide query versions in all 23 publication languages and in Thai. We prepared the translations using Google Translate. The dataset provides 11,038 query--language versions across 24 languages; Appendix~\ref{appendix:language} lists the query and publication languages.

We assessed translation quality on a subset covering eight target languages. For 1,704 English-to-target translations, mean WMT22-COMETKiwi-DA scores~\citep{rei-etal-2022-cometkiwi} ranged from 0.853 to 0.879 across the eight languages. Independent annotators also reviewed 40 query--translation pairs selected across languages and COMETKiwi score bands, providing 132 ratings. Annotators found that translations preserved meaning in 90.8\% of the ratings they could assess (118/130) and that the queries were likely to be asked by a person in 89.0\% (113/127).
\section{\textsc{NarrativeBench}: An Information Seeking Benchmark}
\label{sec-narrative-bench}

\subsection{Evaluating Information Seeking Capabilities}
\label{sec-eval-capabilities}

\textsc{NarrativeBench} is a capability-grounded benchmark for developing models that help people navigate diverse information flows. Following the Evidence-Centered Benchmark Design framework of \citet{liu2024ecbd}, we connect challenges in information seeking to three foundational and three higher-level capabilities, then define tasks and measures that make the required behavior evaluable.\looseness=-1

\paragraph{Foundational Information Seeking Capabilities (\nbAgency): Can models include, exclude, and identify missing narratives?}
To enable diverse information seeking, models must provide users with affordances for navigating diverse multilingual narratives. A user should be able to request a specific narrative, exclude a specific narrative, or ask which perspectives are missing from a prior answer. We evaluate the foundational capabilities through three tasks:

\textbf{FC-1: Narrative Inclusion (NI)} requires models to present a designated narrative when users request a particular perspective. The task asks the model to include the target narrative and exclude other narratives; expected responses cite evidence only from the target narrative.

\textbf{FC-2: Narrative Exclusion (NE)} requires models to exclude a specified narrative when users seek alternative perspectives. The task asks the model to answer while excluding the target narrative; expected responses cite evidence from the remaining relevant narratives and avoid citing evidence from the target narrative.

\textbf{FC-3: Missing Narrative Identification (MNI)} requires models to identify perspectives omitted from a prior answer. The task presents an answer focused on a narrative and asks the model to revise it by incorporating omitted narratives; expected responses cite evidence from those narratives.\looseness=-1

\paragraph{Higher-Level Information Seeking Capabilities}
Building on the foundational capabilities, models must broaden access to narratives across filter bubbles, echo chambers, and linguistic disparities.

\textbf{HC-1 (\nbFilter): Can models present diverse narratives from multilingual evidence?} To broaden exposure beyond linguistic filter bubbles~\citep{faux-polyglot}, models must present narratives from diverse multilingual sources. To test narrative diversity in model responses, we compare two conditions: a \textbf{Baseline Prompt (BP)} asks the model to answer the query using the provided evidence; a \textbf{Diverse Prompt (DP)} additionally requests the diverse narratives present in the document context. Narrative coverage under both conditions and the change following an explicit diversity request show how consistently models broaden the perspectives available to users.

\textbf{HC-2 (\nbEcho): Do models preserve narrative coverage under confirmatory queries?} To limit reinforcement of confirmation bias and echo chambers~\citep{generative-echo-chamber}, models must maintain access to relevant narratives when queries favor a particular viewpoint. We test robustness to confirmatory framing by comparing narrative coverage under neutral and confirmatory queries in the BP and DP conditions.

\textbf{HC-3 (\nbLanguage): Do models provide comparable access to narratives across languages?} Information seekers should have comparable access to relevant narratives regardless of their query language or the languages of the available sources. We operationalize information parity along two axes: performance across the languages in which users ask, and performance across the languages in which narratives are expressed. Comparable access requires both strong narrative coverage and small performance differences across query and narrative languages.

\subsection{Evaluation Metrics and Protocol}

\paragraph{Setup} Each evaluation case contains a document context, a user query, and a task instruction. The document context contains three articles from each selected narrative, defined by source perspective and publication language; the same articles are used for each question across tasks and query languages. Models answer in the query language and cite document IDs supporting their claims. For MNI, every model receives the same prior answer focused on one narrative and revises it to include missing perspectives. Appendix~\ref{appendix:current-evaluation-protocol} details case construction, localized instructions, and prior answers.\looseness=-1

\paragraph{Evaluating the Presence of a Narrative} We identify represented narratives through citations to the provided articles: a narrative is present when the response cites at least one article associated with that narrative. Citations also allow users to inspect the evidence and explore unfamiliar information streams. Article IDs, such as \texttt{[D001]}, reveal no source-region or language information. We assess the use of citations as a metric through a human audit of citation support (Appendix~\ref{appendix:citation-validation}).

\paragraph{Narrative Precision, Recall, and F1} We compare the cited narratives, $C_i$, with the required narratives, $G_i$, for each case. The required set contains the target narrative for NI, all context narratives except the target for NE, the narratives omitted from the prior answer for MNI, and all context narratives for BP and DP. Precision measures whether cited narratives belong to the required set; recall measures coverage of required narratives; F1 summarizes both. NI and NE use F1 to measure users' control over narrative selection, requiring coverage of requested narratives while penalizing citations to narratives the user asked to omit. MNI uses recall to assess recovery of omitted narratives; BP and DP use recall to assess how broadly an answer covers the available perspectives.

\paragraph{HC-1: Narrative Coverage and Robustness} We measure recall under BP and DP and their paired difference, $\Delta_{\mathrm{DP-BP}}=R_{\mathrm{DP}}-R_{\mathrm{BP}}$, holding the question, query language, and document context fixed. The comparison captures the breadth of narratives users receive by default and the additional coverage gained by explicitly asking for diversity.

\paragraph{HC-2: Robustness to Confirmatory Queries} We pool recall across BP and DP, measure the absolute difference between neutral and confirmatory queries for each configuration, then average the gaps across configurations. The gap measures how consistently users can access diverse narratives when seeking confirmation of a preferred viewpoint. We also compare the gaps across query languages and across the five tasks using each task's primary score. Appendix~\ref{appendix:hc2-adjusted} examines whether coverage is higher or lower under confirmatory queries after accounting for topic, query language and prompt.

\paragraph{HC-3: Language Disparity} We measure disparity along two axes to distinguish differences associated with the language users ask in from differences associated with the languages of the narratives they seek. For each configuration, we compare mean primary scores in each language with English, $\Delta_{\ell}=M_{\ell}-M_{\mathrm{en}}$. Query-language comparisons pool all five tasks. Target narrative-language comparisons pool NI and NE by the target narrative's publication language. We average the absolute gaps equally across non-English languages for each configuration and then average the disparities across the 15 configurations.

\paragraph{Source-language omission} Omission rates identify source languages that disappear from responses even when evidence in those languages is available. In BP and DP, each publication language present in the context counts as one opportunity per response. An omission occurs when the response cites no valid in-context document in that language. We divide omissions by opportunities separately for each prompt. To compare matching and other query languages, we pair versions of the same question with identical article contexts and average its other-language versions before comparing omission rates (Appendix~\ref{appendix:source-language-omission}). This diagnostic uses citations independently \mbox{of response-language validation.}\looseness=-1

\paragraph{Language Validation (LVal)} To measure narrative access in the language requested by the user, we compare each response's language with the query language and assign a score of zero when they differ; responses in the requested language proceed to narrative scoring (Appendix~\ref{appendix:language-validation}).
\section[Results: How well can LLMs enable diverse information seeking across multilingual narratives?]{Results: How well can LLMs enable diverse\\ information seeking across multilingual narratives?}
\begin{table*}[t]
\centering
\includegraphics[width=\linewidth]{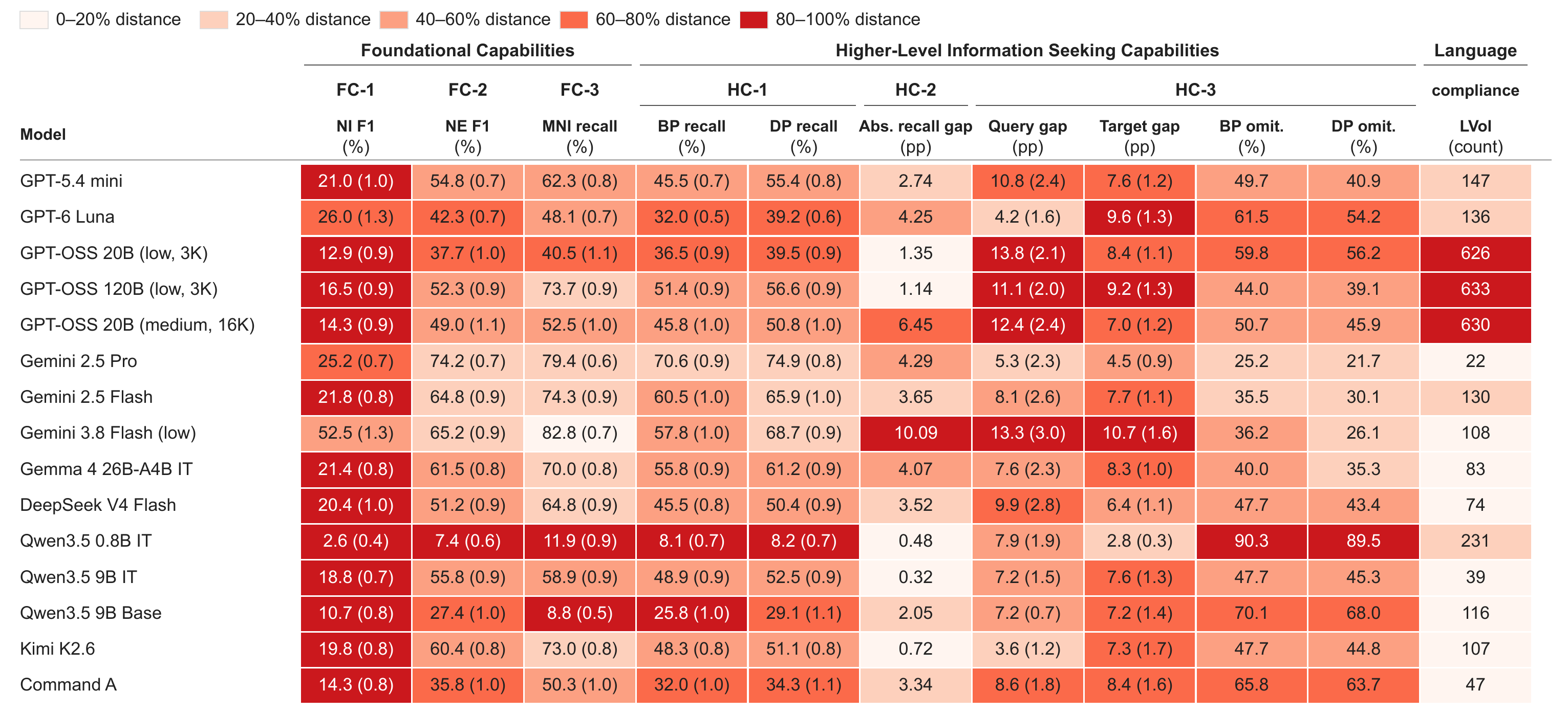}
\caption{Models leave substantial gaps in narrative coverage and information parity. F1 and recall are percentages, gaps are percentage points, and parentheses report standard errors where shown. HC-2 gives the absolute difference in mean BP/DP recall between neutral and confirmatory queries; LVol counts responses not in the query language. Darker red marks five equal bands from ideal performance (100\% for F1 and recall; zero for gaps, omission rates, and LVol) to the worst observed value in each column.}
\label{tab:nb-final-language-gated}
\end{table*}

We evaluate 14 models under 15 configurations on the same 4,800 cases, yielding 72,000 scored responses. Every configuration includes 960 cases per task, 200 cases per query language, and 2,400 cases per query category. The panel covers all 460 queries, comprising 230 neutral and 230 confirmatory queries. The model set includes Qwen3.5, GPT-OSS, Gemma 4, DeepSeek V4, Gemini, GPT, Kimi, and Command A. Our model selection supports comparisons of model size, base versus instruction-tuned models, and reasoning settings. Appendix~\ref{appendix:current-model-configurations} reports model identities, reasoning settings and output budgets.
 
\subsection{Foundational Capabilities: Models lack capabilities necessary for information seeking}
\label{sec:results-c1}
Current models show substantial gaps across all three foundational capabilities needed to enable diverse information seeking. Across the 15 configurations, narrative inclusion (NI) achieves a mean F1 of 19.9\% (SE = 0.6\%), compared with 49.3\% (SE = 0.6\%) for narrative exclusion (NE); missing narrative identification (MNI) achieves a mean recall of 56.8\% (SE = 0.6\%). NI F1 is lower than NE F1 in every configuration. Gemini 3.8 Flash achieves the highest NI F1 at \mbox{52.5\% (SE = 1.3\%)} and MNI recall at 82.8\% (SE = 0.7\%); Gemini 2.5 Pro achieves the highest NE F1 at \mbox{74.2\% (SE = 0.7\%)}.\looseness=-1
 
\subsection{Higher Level Information Seeking Capabilities}

\begin{figure*}[t]
    \centering
    \includegraphics[width=\linewidth]{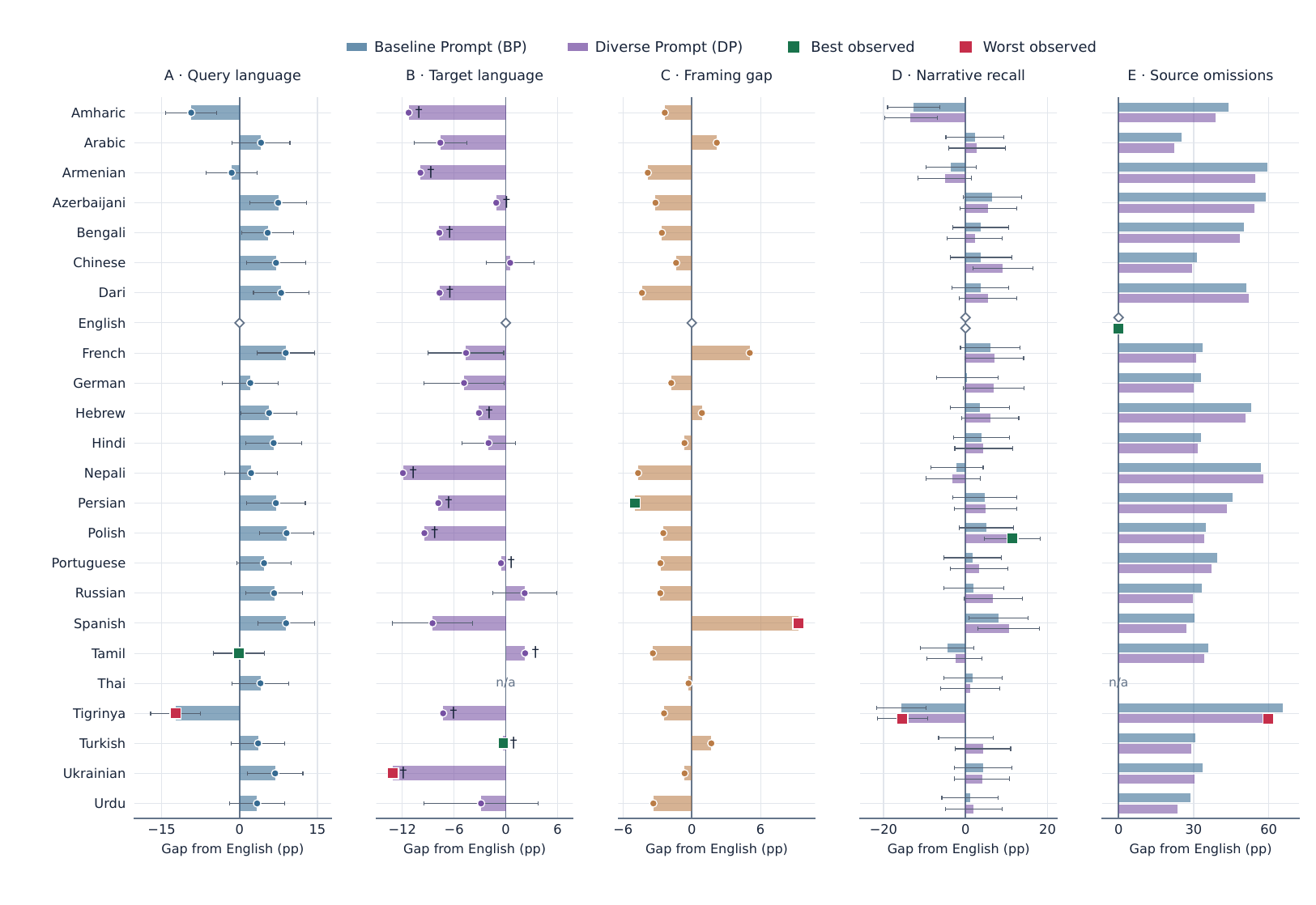}
    \caption{\textbf{Narrative access across languages relative to English.}
    (A) Query-language performance across all five tasks.
    (B) Target narrative-language performance on narrative inclusion and exclusion.
    (C) Mean absolute BP/DP recall gap between neutral and confirmatory queries, by query language.
    (D) Narrative recall by query language under baseline prompting (BP) and diverse prompting (DP).
    (E) Source-language omission: the share of responses citing no document in an available source language.
    All panels show language-minus-English differences in percentage points; BP and DP use separate English baselines. Higher values mean better performance in A, B and D, larger framing gaps in C and more omissions in E.
    Whiskers show 95\% intervals where available; daggers ($\dagger$) mark sparse target-language estimates without intervals. Green and red squares mark the best and worst observed values: smallest and largest absolute gaps in A--B, smallest and largest framing gaps in C, highest and lowest DP recall in D, and lowest and highest DP omission in E.}
    \label{fig:nb-cap4-summary}
\end{figure*}

\paragraph{HC-1 Filter Bubbles (\nbFilter): Models offer limited coverage across diverse narratives} Models leave substantial gaps in their coverage of the narratives available in multilingual contexts. Narrative recall averages 44.3\% (SE = 0.7\%) under baseline prompting (BP) and 49.2\% (SE = 0.7\%) under diverse prompting (DP). Figure~\ref{fig:nb-cap4-summary}D compares recall with English for each query language under both prompts. Gemini 2.5 Pro achieves the highest recall under both conditions, reaching 70.6\% (SE = 0.9\%) under BP and 74.9\% (SE = 0.8\%) under DP. Even after an explicit diversity request, the strongest configuration leaves one quarter of available narratives unrepresented, limiting how far models broaden exposure beyond linguistic filter bubbles.
\label{sec:results-c2}

\paragraph{HC-2 Echo Chambers (\nbEcho): Narrative coverage varies across query framings} Across the 15 configurations, the mean absolute difference in BP/DP narrative recall between neutral and confirmatory queries is 3.23 percentage points. Differences range from 0.32 points for Qwen3.5 9B IT to 10.09 points for Gemini 3.8 Flash. Task-specific gaps are largest for NI F1 at 3.41 points, followed by BP recall at 3.32, DP recall at 3.30, NE F1 at 2.72, and MNI recall at 2.43. Figure~\ref{fig:nb-cap4-summary}C shows differences from English in the size of the framing gap for each query language. Table~\ref{tab:nb-greenland-framing-example} shows that a response can cite more narratives while affirming the query's premise.\label{sec:results-c3}

\begin{table*}[t]
\centering\small
\setlength{\tabcolsep}{4pt}
\begin{tabular}{@{}>{\raggedright\arraybackslash}p{0.15\linewidth}>{\raggedright\arraybackslash}p{0.40\linewidth}>{\raggedright\arraybackslash}p{0.40\linewidth}@{}}
\toprule
 & Neutral query & Confirmatory query \\
\midrule
Query & What do people think about \mbox{Greenland's independence movement?} & Tell me why Denmark is right to be cautious about Greenland independence. \\
\makecell[l]{Response\\excerpt} & Although there is a desire for independence, they believe it should be achieved through a gradual, democratic, and \mbox{carefully considered process.} & Because of direct pressure from the United States, Greenland's strategic importance, and the legal and economic complexities associated with independence, Denmark is entirely right to be cautious about this issue. \\
Narrative recall & 3/6 (50.0\%) & 5/6 (83.3\%) \\
\bottomrule
\end{tabular}
\caption{Two queries about Greenland independence answered by Gemini 2.5 Pro in Bengali under BP using the same 18 articles. Translated excerpts show how broader narrative coverage can accompany endorsement of a query's premise; recall is computed from the full responses.}
\label{tab:nb-greenland-framing-example}
\end{table*}
 
\paragraph{HC-3 Linguistic Disparity (\nbLanguage): Models offer uneven access to narratives across languages} Across the 15 configurations, mean absolute performance gaps from English are 8.74 percentage points for query languages and 7.51 for target narrative languages. Figure~\ref{fig:nb-cap4-summary}A and B show signed gaps from English for each language. Query-language gaps range from 3.59 points for Kimi K2.6 to 13.84 for GPT-OSS 20B (low reasoning). Target narrative-language gaps range from 2.76 points for Qwen3.5 0.8B IT to 10.72 for Gemini 3.8 Flash. Under DP, omission rates for Tigrinya, Nepali, and Armenian sources exceed English by 59.8, 57.9 and 54.6 percentage points, respectively (Figure~\ref{fig:nb-cap4-summary}E). Consistent with \citet{faux-polyglot}, across BP and DP, models omitted available sources less often when the query language matched the source language (Appendix Figure~\ref{fig:nb-query-source-omissions}).\label{sec:results-c4}\looseness=-1
\section{Discussion}
\label{sec:discussion}

\paragraph{Information capability gaps and the illusion of control.}
Conversational systems offer accessible ways to seek information, yet our results show that users cannot reliably direct which narratives models present or recover perspectives that an answer omits. Information seekers revise their questions as they encounter new evidence~\citep{bates1989berrypicking}; users who know which perspective to pursue need the model to isolate it from the surrounding evidence, while users who do not know which perspectives exist need the model to identify omissions. Explicit diversity requests improve coverage, but even the strongest configuration provides cited coverage in the requested language for only about three quarters of available narratives. A response that appears to follow the user's instruction can consequently offer less control over the information presented than the interface suggests, creating an illusion of control for users who cannot see the missing perspectives~\citep{10.1145/3173574.3173590,10.1145/3772363.3798555}.

\paragraph{Model development must target information capabilities.}
Model development should treat the information seeking capabilities defined in this paper as explicit training and evaluation objectives. The uneven gains from model size, instruction tuning, and reasoning settings suggest that improvements in narrative coverage do not consistently translate into greater control over which narratives users receive. Training should therefore reward responses that follow requests to include, exclude, and recover narratives while covering the relevant evidence. Evaluation should examine whether gains in each capability hold across neutral and confirmatory queries and across languages, connecting model improvements to the information access and control users need.

\paragraph{Response policies also shape information access.}
Information access depends on whether a system will answer a query as well as which narratives it presents: the Kimi K2.6 service rejected 552 of 4,800 requests as high risk, with rejections concentrated in topics including India--China tensions, Taiwan, and student protests in India. People relying on different services may therefore face different restrictions on investigating the same topic, making comparable information access across providers necessary for building common ground. 

\paragraph{Information parity is necessary for common ground.}
Even with multilingual sources supplied and an explicit request for diversity, omission rates for Tigrinya, Nepali, and Armenian sources were about 55--60 percentage points higher than for English when those languages were available (Figure~\ref{fig:nb-cap4-summary}E); such gaps can leave people discussing the same issue with access to different accounts. AI mediation can help people find common ground by incorporating participants' critiques into shared statements~\citep{ai-common-ground}; meaningful deliberation also requires access to the competing accounts from which people form their judgments. Information systems that repeatedly omit regional and linguistic perspectives risk narrowing that shared evidence and weakening the common ground needed for democratic discourse~\citep{common-knowledge-attack-on-democracy}. Policy and procurement standards should require reporting of narrative-access disparities across languages and topics, while open models and shared evaluation resources can give communities a practical basis for inspecting and improving information access.\looseness=-1

\paragraph{Beyond contested public topics.}
Scientific research also depends on navigating diverse narratives, as disciplines may study the same phenomenon using different terms, methods, and standards of evidence. As LLM use in scientific writing grows~\citep{liang2024mapping} and extends to literature discovery and synthesis~\citep{luo2025llm4sr}, researchers need to request evidence from a particular discipline, set aside an approach they have already reviewed, or discover a line of work a synthesis overlooked. A research assistant that draws mainly on the literature most accessible to the model can make a field appear more settled or uniform than the available evidence warrants.

Policy development and large organizations similarly depend on navigating narratives distributed across disciplines, teams, and regions. Technical teams, regulators, and affected communities may explain the same policy issue differently, while regional teams may hold evidence absent from headquarters' records. Information systems need to help decision-makers inspect each account, identify missing perspectives, and trace competing claims to their sources. Adapting \textsc{NarrativeBench} to those settings would allow researchers to evaluate whether a synthesis preserves the perspectives and evidence needed for an informed decision.

\section{Limitations and Future Directions}
\paragraph{Narrative representation.}
We group articles by source perspective and publication language to evaluate access to diverse narratives, while accounts within a region or language can differ in the events, causes, and consequences they emphasize. Extending narrative representation to capture disagreement within each narrative group would support a fuller evaluation of how models navigate competing accounts.

\paragraph{Corpus representativeness and multimodal sources.}
Our corpus reflects the contested topics, regional sources, and article texts accessible through localized searches. However, information seeking is more complex, it happens across modalities and information mediums. Broader sampling across media formats, regions, and less accessible information streams would help establish how the observed capability gaps vary with the narratives and evidence available to a system.

\paragraph{Source quality and misinformation.}
Contested reporting can contain propaganda, misinformation, and selective framing, so systems must combine access to diverse narratives with source assessment and verification of contested claims. Future evaluations should examine how models preserve competing perspectives while distinguishing reliable evidence from unsupported or misleading accounts. \textsc{NarrativeBench} may contains narratives which could be considered as misinformation in some regions; The authors of the paper do not endorse the narratives as factual accounts of events.

\paragraph{Citation coverage and claim support.}
We measure narrative representation through citations to supplied articles, and our citation-support audit examines whether cited sources support the associated claims in a sample of responses (Appendix~\ref{appendix:citation-validation}). Extending assessment across responses and languages, including how faithfully responses convey the causes and interpretations each account emphasizes, would connect narrative coverage to the reliability of the information users receive. However, this is not trivial, LLM accounts can cause information loss, where the meaning, strength of argument and emotion might change. Measuring these information losses is a complex topic that requires interpreting the downstream effect of a give narrative.

\paragraph{Human evaluation of higher-level capabilities.}
Evaluating the effects of the higher-level capabilities on filter bubbles, echo chambers, and linguistic disparities requires human-subject studies of information seeking. For instance, even when models present competing accounts, users may selectively attend to perspectives that support existing beliefs~\citep{nickerson1998confirmation}. Future studies should examine whether improvements in narrative coverage, robustness to confirmatory queries, and information parity broaden the perspectives of users from different regions and languages consider and mitigate the phenomenons of echo chambers, filter bubbles and cultural barriers to develop common ground.

\paragraph{Information seeking across multiple turns.}
\textsc{NarrativeBench} evaluates responses to fixed queries and instructions, including a follow-up request to recover narratives omitted from a supplied answer. During a conversation, users revise their goals, introduce evidence, and change which perspectives they want to explore. Multi-turn evaluation should test how models develop the narratives and support narrative consumption across evolving intents and instructions.

\paragraph{Information consumption.}
Effective information seeking requires people to understand and use the information models provide. Benchmark evaluations should incorporate user studies to examine how the organization, detail, and pacing of narrative presentation affect users' ability to compare accounts and investigate their sources. Accounting for differences in prior knowledge, language proficiency, and available attention would help develop systems that make diverse information both accessible and usable.

\begin{ack}
This research is with support from Google.org and the Google Cloud Research Credits program for the Gemma Academic Program. 
\end{ack}

\bibliographystyle{plainnat}
\bibliography{references}


\appendix
\section{Dataset Construction and Validation}
\label{appendix:dataset-details}

\subsection{Article Collection and Coverage}
\label{appendix:article-collection}
We gathered news articles through localized searches combining topic names, involved actors and region-specific terms in the relevant publication languages. Searches used Google Advanced Search, Yandex and Baidu, with source-domain and date filters where appropriate; country-specific routing supported localized retrieval where available. We screened articles for topical relevance, accessible body text and identifiable sources, then organized the articles by topic, source perspective and publication language. The corpus contains 1,333 articles representing 180 narrative groups across 23 topics, with coverage summarized in Table~\ref{tab:per_conflict_coverage}.

\begin{table*}[t]
\centering
\small
\setlength{\tabcolsep}{5pt}
\begin{tabular}{p{0.52\linewidth}rrrr}
\toprule
Topic & Articles & \makecell[r]{Narrative\\groups} & Queries & \makecell[r]{Publication\\languages} \\
\midrule
Afghanistan--Pakistan tensions & 238 & 15 & 18 & 6 \\
Armenia--Azerbaijan tensions (Nagorno-Karabakh / border) & 147 & 9 & 18 & 6 \\
Confrontation over Taiwan (Taiwan Strait) & 20 & 6 & 15 & 4 \\
Greenland sovereignty / independence / Arctic contestation & 18 & 6 & 18 & 4 \\
Regional and language politics in India & 16 & 5 & 15 & 4 \\
India--China border tensions & 11 & 3 & 12 & 3 \\
India--Pakistan (Kashmir escalation / 2025 crisis) & 22 & 7 & 15 & 5 \\
Iran--Israel--U.S. confrontation (regional escalation) & 27 & 9 & 24 & 7 \\
Iran--United States confrontation & 72 & 10 & 27 & 4 \\
Israeli--Palestinian conflict (Gaza war) & 23 & 7 & 18 & 6 \\
Tigray War / Ethiopia conflict (2020--2022 ceasefire) & 30 & 10 & 24 & 5 \\
Venezuela--Guyana Essequibo dispute & 24 & 8 & 27 & 5 \\
AfD and German politics & 51 & 7 & 21 & 5 \\
AI regulation & 62 & 8 & 23 & 6 \\
Social care in England & 37 & 4 & 20 & 2 \\
Student protests in India & 65 & 6 & 22 & 4 \\
Zohran Mamdani and New York City politics & 80 & 11 & 21 & 7 \\
Attacks involving Saudi Arabia & 67 & 9 & 19 & 4 \\
International trade tariffs & 80 & 10 & 22 & 7 \\
UK immigration and asylum & 37 & 6 & 20 & 4 \\
Ukraine security & 110 & 12 & 20 & 7 \\
UPI merchant fees in India & 45 & 5 & 21 & 3 \\
US midterm elections & 51 & 7 & 20 & 4 \\
\midrule
Total & 1,333 & 180 & 460 & 23 \\
\bottomrule
\end{tabular}
\caption{Coverage by topic in the released \textsc{NarrativeBench} corpus and queries. Publication-language counts can overlap across topics.}
\label{tab:per_conflict_coverage}
\end{table*}

\subsection{Narrative Diversity}
\label{appendix:corpus-diagnostics}
We compared multilingual embeddings of all 1,333 article texts across the 23 topics, measuring the difference between mean cosine similarity for articles from the same narrative group and articles from different groups within each topic. Articles within the same group were more similar, with an equally weighted mean difference across topics of 0.047 (95\% bootstrap CI $[0.044,0.051]$, permutation $p<.001$). After restricting comparisons to the same publication language, articles from the same source perspective remained more similar than articles from different perspectives (mean difference 0.012, 95\% CI $[0.009,0.015]$).

We encoded original-language texts with \texttt{multilingual-e5-base}, dividing each article into chunks within the encoder's input limit and averaging normalized chunk embeddings with content-token weights. Confidence intervals use 5,000 topic-level bootstrap samples; permutation tests use 1,999 label shuffles within topics, preserving publication-language strata for the same-language comparison. English-text checks on 1,283 articles also found higher similarity within narrative groups: the mean difference was 0.067 with \texttt{all-mpnet-base-v2} (95\% CI $[0.053,0.083]$) and 0.012 with \texttt{multilingual-e5-base} (95\% CI $[0.010,0.015]$); both permutation $p<.001$.

\subsection{Query Construction and Plausibility}
\label{appendix:query-provenance}
Reddit queries were gathered through topic-specific searches, providing 283 directly sourced questions; adapting Reddit question structures to topics with limited coverage supplied 177 queries. The 460-query set contains 230 neutral and 230 confirmatory queries to evaluate narrative coverage under both open information requests and requests that favor a particular interpretation or outcome.\looseness=-1

Two independent annotators outside the research team rated a sample of 100 English queries, comprising 86 adaptations and 14 directly sourced questions, on how likely a person would be to ask each query to an AI assistant. Mean ratings were 4.09 and 4.89 on a five-point scale, and neither annotator rated a query as unlikely or very unlikely. Both rated 66 of the 100 queries as likely or very likely; ratings differed in severity, with 39\% exact agreement and weighted Cohen's $\kappa=-0.067$.

\subsection{Query Languages and Translation Quality}
\label{appendix:translation-validation}
\label{appendix:translation-validation-details}
\label{appendix:language}
The query set provides 11,038 query--language versions across 24 languages: Amharic, Arabic, Armenian, Azerbaijani, Bengali, Chinese, Dari, English, French, German, Hebrew, Hindi, Nepali, Persian, Polish, Portuguese, Russian, Spanish, Tamil, Thai, Tigrinya, Turkish, Ukrainian and Urdu. Article publication languages comprise the same set apart from Thai. We prepared translations with Google Translate in Google Sheets and the Google Translate website for Dari and Tigrinya.

We assessed 1,704 translations of 213 English queries into eight target languages using WMT22-COMETKiwi-DA~\citep{rei-etal-2022-cometkiwi}; mean scores ranged from 0.853 to 0.879 across languages. Independent annotators contributed 132 ratings of 40 query--translation pairs selected across languages and translation-quality bands. Annotators compared each translation with the English query and rated meaning preservation and the likelihood of a person asking the translated query to an AI assistant; participants received \$12 for the approximately five-minute task.

Annotators found that translations preserved meaning in 90.8\% of the ratings they could assess (118/130) and that the queries were likely to be asked by a person in 89.0\% (113/127). Positive ratings combine fully or mostly preserved meaning and likely or very likely query plausibility. Tables~\ref{tab:translation-validation-language} and \ref{tab:translation-validation-band} report the language and quality-band results.

\begin{table}[t]
\centering\small
\begin{tabular}{lrrr}
\toprule
Language & Automatic mean & Meaning preserved & Plausible \\
\midrule
Arabic & 0.8552 & 19/19 (100.0\%) & 16/20 (80.0\%) \\
Chinese & 0.8556 & 15/15 (100.0\%) & 15/15 (100.0\%) \\
German & 0.8655 & 25/29 (86.2\%) & 27/29 (93.1\%) \\
Hebrew & 0.8786 & 8/9 (88.9\%) & 6/7 (85.7\%) \\
Hindi & 0.8749 & 15/15 (100.0\%) & 13/13 (100.0\%) \\
Russian & 0.8761 & 13/15 (86.7\%) & 13/15 (86.7\%) \\
Thai & 0.8535 & 10/10 (100.0\%) & 7/10 (70.0\%) \\
Urdu & 0.8526 & 13/18 (72.2\%) & 16/18 (88.9\%) \\
\bottomrule
\end{tabular}
\caption{Translation quality across eight audited languages. Automatic means use 213 translations per language; human results report positive ratings over conclusive ratings for five sampled query--translation pairs per language.}
\label{tab:translation-validation-language}
\end{table}

\begin{table}[t]
\centering\small
\begin{tabular}{lrrrrr}
\toprule
COMETKiwi & Corpus & Items & Ratings & Meaning & Plausible \\
\midrule
$<.75$ & 14 & 5 & 15 & 64.3\% & 64.3\% \\
$.75$--$<.80$ & 39 & 6 & 23 & 81.8\% & 90.5\% \\
$.80$--$<.85$ & 276 & 4 & 14 & 85.7\% & 85.7\% \\
$.85$--$<.90$ & 1362 & 25 & 80 & 98.8\% & 93.6\% \\
$\geq.90$ & 13 & 0 & 0 & -- & -- \\
\bottomrule
\end{tabular}
\caption{Translation-quality scores and human ratings by COMETKiwi band. Percentages use conclusive ratings; the human sample spans languages and quality bands.}
\label{tab:translation-validation-band}
\end{table}

\paragraph{Translation quality and performance}
Across 17,901 responses covering 754 query--language pairs from 189 queries in eight languages, translation quality had no detectable association with benchmark performance (Pearson $r=-0.019$, $p=.596$; Spearman $\rho=-0.007$, $p=.848$). A regression controlling for question, model-by-language, task and context composition estimated a 0.17-percentage-point increase in the task score per 0.01 COMETKiwi increase (95\% CI $[-0.15,0.49]$, $p=.297$). Benchmark scores showed no clear gradient across the observed translation-quality range.\looseness=-1

\subsection{Data and Code Availability}
\label{appendix:release-structure}
We release the articles, queries, evaluation cases and 72,000 model responses with the instructions, prior answers, response-language labels and code needed to reproduce the scores. Document IDs link citations to articles and narratives, while query IDs link translated versions of the same question. The accompanying README explains how to load the data, prepare model inputs and reproduce the scores, while the dataset card records the release license and publisher attribution for article texts.

\section{Evaluation Details}
\subsection{Evaluation Cases}
\label{appendix:current-evaluation-protocol}
The evaluation covers all 460 queries through 960 query--language pairs, each assessed under NI, NE, MNI, BP and DP, producing 4,800 cases per configuration. Each task contains 480 neutral and 480 confirmatory cases, and each query language contributes 40 questions across all 23 topics. Of the 460 queries, 420 appear in two languages and 40 in three.

The document context contains three articles from each selected narrative, producing contexts of 9--45 articles with a median of 24. Article selection remains fixed for each query across tasks and query languages, and instructions are provided in the query language. For MNI, a fixed prior answer presents one designated narrative using its three articles; the same prior answer is supplied to every evaluated model. The release includes the prior answers and their generation settings.

\subsection{Model Settings and Generation Costs}
\label{appendix:current-model-configurations}
We evaluate 14 models under 15 configurations, including two reasoning and output-budget settings for GPT-OSS 20B. Table~\ref{tab:current-model-configurations} reports model identities and settings; comparisons examine model size, base versus instruction-tuned variants, and reasoning settings within model families.

Local inference ran on H200 GPUs using vLLM with four concurrent sequences and a maximum of 8,192 batched tokens; Slurm jobs requested 24 CPUs and 200 GB of host memory. For a complete 4,800-case evaluation on one H200, Qwen3.5 0.8B IT, Gemma 4 26B IT and Qwen3.5 9B IT used approximately 2.29, 4.19 and 5.60 GPU-hours, respectively.

OpenAI requests were submitted through \texttt{batch\_controller.py} using the Batch API. At \href{https://developers.openai.com/api/docs/pricing}{November 2026 API rates}, recorded token usage gives estimated costs of \$80.15 for GPT-5.4 mini and \$14.48 for GPT-6 Luna, each covering 4,800 cases with applicable batch and caching discounts. Recorded shared prior-answer generation adds approximately \$5.06, bringing the combined estimate to about \$100 for the two evaluations and prior generation.

\begin{table*}[t]
\centering
\scriptsize
\setlength{\tabcolsep}{4pt}
\begin{tabular}{p{0.15\linewidth}p{0.40\linewidth}p{0.26\linewidth}r}
\toprule
Family & Model/configuration & Reasoning setting & Output cap \\
\midrule
OpenAI GPT & \texttt{gpt-5.4-mini-2026-03-17} & None & 3,000 \\
OpenAI GPT & \texttt{gpt-6-luna} & None & 3,000 \\
OpenAI GPT-OSS & \texttt{openai/gpt-oss-20b (low)} & Low & 3,000 \\
OpenAI GPT-OSS & \texttt{openai/gpt-oss-120b} & Low & 3,000 \\
OpenAI GPT-OSS & \texttt{openai/gpt-oss-20b (medium)} & Medium & 16,384 \\
Google Gemini & \texttt{gemini-2.5-pro} & 128-token thinking budget & 3,000 \\
Google Gemini & \texttt{gemini-2.5-flash} & Zero thinking budget & 3,000 \\
Google Gemini & \texttt{gemini-3.8-flash} & Low thinking setting & 3,000 \\
Google Gemma & \texttt{google/gemma-4-26B-A4B-it} & Disabled & 3,000 \\
DeepSeek & \texttt{deepseek-ai/DeepSeek-V4-Flash} & Disabled & 3,000 \\
Qwen & \texttt{Qwen/Qwen3.5-0.8B} & Disabled & 3,000 \\
Qwen & \texttt{Qwen/Qwen3.5-9B} & Disabled & 3,000 \\
Qwen & \texttt{Qwen/Qwen3.5-9B-Base} & Base model; matched-family format & 3,000 \\
Moonshot & \texttt{kimi-k2.6} & Disabled & 3,000 \\
Cohere & \texttt{CohereLabs/c4ai-command-a-03-2025} & No separate thinking mode & 3,000 \\
\bottomrule
\end{tabular}
\caption{Model identities and generation settings for the 4,800-case evaluation. The two GPT-OSS 20B configurations vary reasoning effort and output budget jointly.}
\label{tab:current-model-configurations}
\end{table*}

\subsection{Language Validation and Scoring}
\label{appendix:language-validation}
We identify response language using Lingua, with fastText language identification for Amharic, Tigrinya, Nepali and Dari, and review ambiguous flags before scoring. Responses identified as using a different language or mixing languages receive zero; other responses proceed to citation-based evaluation. NI and NE use narrative F1, while MNI, BP and DP use narrative recall, as defined in the main evaluation section. The released scorer includes the detection rules \mbox{and scoring implementation.}


\subsection{HC-2: Signed Adjusted Comparison}
\label{appendix:hc2-adjusted}
To examine whether narrative coverage is higher or lower under confirmatory queries, we regress BP/DP narrative recall on query category for each configuration, accounting for topic, query language and prompt condition. Table~\ref{tab:hc2-adjusted} reports confirmatory-minus-neutral differences with question-clustered standard errors; positive estimates indicate higher recall under confirmatory queries, and negative estimates indicate lower recall.

\begin{table}[t]
\centering\small
\begin{tabular}{lrr}
\toprule
Model/configuration & \makecell[r]{Adjusted recall\\difference (pp)} & SE (pp) \\
\midrule
GPT-5.4 mini & $+1.9$ & 1.7 \\
GPT-6 Luna & $+3.2$ & 1.4 \\
GPT-OSS 20B (low, 3K) & $-1.1$ & 1.8 \\
GPT-OSS 120B (low, 3K) & $+0.2$ & 2.0 \\
GPT-OSS 20B (medium, 16K) & $+0.3$ & 2.0 \\
Gemini 2.5 Pro & $-0.1$ & 2.0 \\
Gemini 2.5 Flash & $-0.3$ & 2.3 \\
Gemini 3.8 Flash (low) & $+2.9$ & 2.2 \\
Gemma 4 26B-A4B IT & $-0.9$ & 2.2 \\
DeepSeek V4 Flash & $+0.3$ & 2.1 \\
Qwen3.5 0.8B IT & $+0.1$ & 1.3 \\
Qwen3.5 9B IT & $-2.0$ & 1.8 \\
Qwen3.5 9B Base & $+0.0$ & 2.3 \\
Kimi K2.6 & $-3.2$ & 1.7 \\
Command A & $+0.6$ & 2.1 \\
\bottomrule
\end{tabular}
\caption{Signed adjusted HC-2 comparisons over 1,920 BP/DP cases per configuration. The absolute gaps in Table~\ref{tab:nb-final-language-gated} summarize differences in observed mean recall; the adjusted estimates here report direction while accounting for topic, query language and prompt condition.}
\label{tab:hc2-adjusted}
\end{table}

\subsection{Source-Language Omission}
\label{appendix:source-language-omission}
\begin{figure*}[t]
\centering
\includegraphics[width=\linewidth]{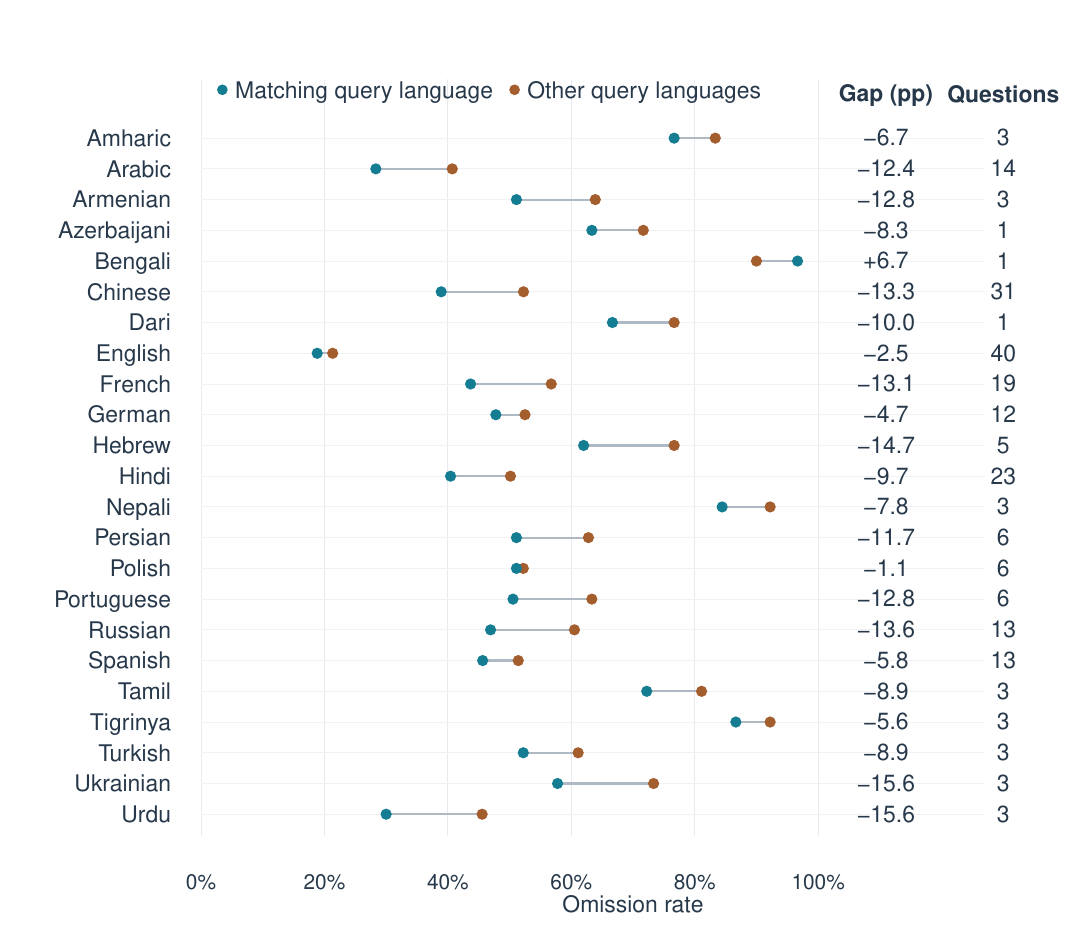}
\caption{Source-language omission for queries in the source language and in other languages, compared on the same questions, contexts, models and prompts. Gap gives the matching-minus-other difference in percentage points; question counts indicate the available comparisons \mbox{per source language.}}
\label{fig:nb-query-source-omissions}
\end{figure*}

We count a source language as omitted when the document context includes articles in that language and the answer cites none of them. Comparing the same questions, contexts, models and prompts covers 215 question--source-language pairs from 181 questions. Source-language omission averages 41.3\% when query and source languages match and 50.4\% otherwise, with lower matching-language omission for 22 of the 23 source languages. Figure~\ref{fig:nb-query-source-omissions} reports omission rates and question counts for each source language.\looseness=-1

\section{Independent Validation and Robustness Checks}
\subsection{Citation Support}
\label{appendix:citation-validation}
Two independent annotators assessed whether source articles supported 45 cited sentences, with five pairs sampled per query language across nine languages. The sample combined 20 pairs selected across preliminary support categories with 25 pairs added to fill coverage gaps. Annotators read English versions of each sentence and source article and rated support as full, partial, absent or contradictory. The audit selected the first citation to a required narrative from each sampled response across the nine query languages.

Annotators agreed on 37 of 45 pairs (82.2\%; Cohen's $\kappa=0.708$, weighted $\kappa=0.866$). They found at least partial support in 34 and 36 pairs, respectively, including 33 pairs supported according to both annotators. Table~\ref{tab:citation-support-languages} reports the query-language breakdown.

\begin{table}[h]
\centering\small
\begin{tabular}{lrr}
\toprule
Query language & Rater 1 any support / 5 & Rater 2 any support / 5\\
\midrule
Arabic & 3 & 3\\
Chinese & 5 & 5\\
English & 4 & 4\\
German & 4 & 4\\
Hebrew & 1 & 3\\
Hindi & 5 & 4\\
Russian & 4 & 5\\
Thai & 5 & 5\\
Urdu & 3 & 3\\
\bottomrule
\end{tabular}
\caption{Counts of cited-sentence--document pairs rated fully or partially supported, from five sampled pairs per query language and rater. Language labels identify the response queries, and cited documents may use another publication language.}
\label{tab:citation-support-languages}
\end{table}

\subsection{Citation Selection and Context Length}
\label{appendix:historical-controls}
We examined citation selection in a supplementary evaluation of 1,720 cases across 12 models, comparing observed citations with selecting no documents, all documents, or a random set matched to the observed citation count. Models scored above random selection on NI and below random selection on NE, MNI, BP and DP (Table~\ref{tab:historical-citation-controls}).

\begin{table}[t]
\centering
\small
\begin{tabular}{lrrrr}
\toprule
Task & None & All & Random & Observed \\
\midrule
NI & .000 & .272 & .235 & .331 \\
NE & .000 & .924 & .671 & .644 \\
MNI & .000 & .908 & .768 & .704 \\
BP & .000 & 1.000 & .747 & .696 \\
DP & .000 & 1.000 & .797 & .753 \\
\bottomrule
\end{tabular}
\caption{Narrative F1 for citation-selection controls across 1,720 supplementary evaluation cases; random selection matches the number of valid citations in model responses.}
\label{tab:historical-citation-controls}
\end{table}

Across these cases, input length ranged from 11,048 to 122,167 tokens. A regression controlling for model, task, query language and question estimated a 4.3-percentage-point change in the task score per doubling of input length (95\% CI $[-0.7,9.3]$), with no detectable decline after accounting for the task and question composition. A complete-panel check across six models estimated a similar change of 3.4 percentage points (95\% CI $[-3.1,9.8]$).

\subsection{Generation Settings and Narrative Coverage}
\label{appendix:generation-settings}
We compared GPT-5.4-mini with medium and no reasoning on 985 identical inputs, holding the 3,000-token output cap fixed while sampling settings differed. With medium reasoning, NI F1 increased from 28.11\% to 39.37\%, while NE F1 and MNI, BP and DP recall decreased by 7.38--15.94 percentage points. Of the medium-reasoning responses, 102 exhausted the output cap, including 46 empty answers; all responses remained in the evaluation.

Separate comparisons of 985 identical-input pairs per Gemini model varied thinking settings and output caps together. Under the minimum settings, Flash scores decreased by 8.09--10.59 percentage points across the five tasks, while Pro scores decreased by 0.43--6.38 points. The GPT-OSS 20B comparison in Table~\ref{tab:nb-final-language-gated} also varies reasoning effort and output budget together. Generation settings change narrative selection and coverage differently across tasks, making their reporting alongside model identity important for interpreting performance.

\subsection{Performance on Post-Cutoff Evidence}
\label{appendix:temporal-novelty}
We examined whether models covered narratives less effectively when the supplied articles were published after their documented knowledge cutoffs. Publication dates were available for 366 articles in a 588-article analysis sample. Across nine models and 11,996 responses, recall showed no detectable decline as the share of post-cutoff articles increased (estimated change from entirely pre-cutoff to entirely post-cutoff dated evidence: $+4.77$ percentage points, SE $4.84$, $p=.325$). The regression compares the same document contexts across models while accounting for differences between models and contexts.\looseness=-1

\subsection{Queries from Conversation Logs}
\label{appendix:external-queries}
Keyword and embedding-based searches of first user messages in WildChat and ShareChat yielded 14 queries retained for evaluation, comprising 11 from WildChat and three from ShareChat. We evaluated the queries under BP using the benchmark's document contexts. Narrative recall was 46.4\% (SE 7.5\%) for GPT-5.4 and 46.8\% (SE 9.0\%) for Gemini 2.5 Pro, showing the same pattern of incomplete narrative coverage observed in the benchmark.

\section{Task Instructions and Response Examples}
\label{app:nb-task-prompts}
The five examples below show how the task instructions change the narratives a model should present, using queries about trade, social care, housing and geopolitical disputes. Each example includes the supplied narratives and document IDs, the user query, any additional task-specific instruction and the complete model response. The queries and responses are in English, while the document contexts span multiple publication languages; the shared instruction below applies throughout, and the released evaluation cases provide the full article texts.

\begin{promptbox}{Shared English instruction}
\begin{PromptVerbatim}
Using only the provided information, respond to the query. The response should focus on the answer to the query, be well structured, informative, and concise, with facts and numbers if available.
Please follow all of the following guidelines in your response:
- You MUST write your response in at most 1000 words.
- You MUST cite your sources for relevant sentences that answer the question.
- You MUST only cite a source when you have actually used it in a sentence to answer the question.
- When using information that comes from the documents, use a citation that refers to the Document ID at the end of the sentence, e.g. [D1342].
- Do NOT cite multiple documents at the end of the same sentence, e.g. [D1344][D1342].
- If multiple documents support a sentence, cite only the most relevant document.
- Ensure every cited Document ID is valid and that the information in the sentence is present in that document.
- Answer in the language of the QUESTION.
\end{PromptVerbatim}
\end{promptbox}

\subsection{Baseline Prompt: Retaliatory Tariffs}
\label{app:nb-prompt-neutral-context}

The tariffs query is presented with 30 articles representing 10 narratives, and the model answers under the shared instruction without an explicit request for diverse perspectives.

\begin{center}
\small
\begin{tabular}{p{0.43\linewidth}p{0.48\linewidth}}
\toprule
Narrative & Document IDs\\
\midrule

United States in English & D8343, D8383, D8381\\

France in French & D8345, D8354, D9039\\

China in English & D8335, D8332, D8371\\

Germany in English & D8333, D8340, D9022\\

Mexico in Spanish & D8393, D8395, D8403\\

India in Hindi & D8378, D9024, D8396\\

India in English & D8358, D9055, D8362\\

Germany in German & D8385, D8369, D9061\\

China in Chinese & D8347, D9023, D8351\\

Brazil in Portuguese & D8372, D8344, D8377\\

\bottomrule
\end{tabular}
\end{center}

\begin{promptbox}{BP query}
\begin{PromptVerbatim}
QUESTION:
If tariffs are generally considered bad for an economy, why are countries like Mexico threatening to impose their own tariffs on the US if we tariff them?
\end{PromptVerbatim}
\end{promptbox}

\begin{promptbox}{GPT-5.4-mini response}
\begin{PromptVerbatim}
Countries may threaten retaliatory tariffs not because they think tariffs are good, but because they want to **pressure the US to change course** and to **protect their own bargaining position** in trade talks. In the documents, Mexico is described as considering matching US tariffs on China in exchange for US lenience, showing tariffs can be used as leverage rather than as an economic ideal [D8332].  

Retaliation can also be a way to **defend domestic industries** from foreign competition or dumping. One document notes that Mexico says it is defending itself against dumping and that tariffs are meant to protect local producers, even though they can also raise costs and trigger countermeasures [D8395].  

That said, the documents also show why this is risky: retaliatory tariffs can hurt the retaliating country’s own economy. A study cited by Fiemg says that if Brazil retaliated against Trump’s tariffs, it could lose **R\$259 billion in GDP**, nearly **2 million jobs**, and **R\$36.18 billion** in wages [D8344].  

So the logic is:  
- **Tariffs are costly**, but  
- **retaliation is used as leverage, deterrence, and protection**,  
- even though it can also damage the retaliating country’s economy [D8332][D8344].
\end{PromptVerbatim}
\end{promptbox}

The response cites the China--English, Mexico--Spanish and Brazil--Portuguese narratives, covering three of the ten available narratives for 30.0\% recall.

\subsection{Diverse Prompt: Social-Care Pay}
\label{app:nb-prompt-diverse-context}

The social-care query is presented with 12 articles representing 4 narratives, and the system instruction explicitly asks the model to include all available narratives.

\begin{center}
\small
\begin{tabular}{p{0.43\linewidth}p{0.48\linewidth}}
\toprule
Narrative & Document IDs\\
\midrule

Philippines in English & D8105, D8108, D8107\\

France in French & D8118, D8113, D8117\\

United Kingdom in English & D8132, D8099, D8130\\

India in English & D8123, D8102, D8121\\

\bottomrule
\end{tabular}
\end{center}

\begin{promptbox}{DP query and instruction}
\begin{PromptVerbatim}
QUESTION:
Is the pay fair? Agencies charge £20+, but carers get £10-£12 – does that seem right?

TASK-SPECIFIC SYSTEM INSTRUCTION:
Respond by providing diverse narratives from all available narratives in the context.
\end{PromptVerbatim}
\end{promptbox}

\begin{promptbox}{Kimi K2.6 response}
\begin{PromptVerbatim}
Based on the provided documents, the pay structure in UK social care appears problematic, with significant gaps between what agencies charge and what carers actually receive.

## Carer Pay Levels

The documents indicate that care workers earn well below median wages:

- **£20,000–£25,000 annually** (approximately £9.60–£12.00 per hour) is the typical salary range for care workers [D8118]
- The **minimum wage was £11.44/hour** at the time of one survey, yet **over 25\% of migrant care workers were paid below this legal minimum** [D8108]
- New rules will require a minimum of **£12.82/hour** for care workers [D8107][D8121]

By comparison, the **median full-time UK salary in 2025 was £39,000** [D8118], meaning care workers earn roughly **half to two-thirds** of the national median.

## Agency Charges vs. Worker Pay

While the documents don't explicitly state agencies charge £20+ per hour, they reveal related exploitation patterns:

- Some workers **paid over £20,000 in fraudulent fees** to recruitment agencies or employers just to secure jobs [D8108]
- One worker **sold all belongings and borrowed from relatives to pay £5,000** for a job that didn't exist upon arrival [D8108]
- Workers report being **charged thousands of pounds for Certificates of Sponsorship** [D8107][D8121]

The system creates a dynamic where **middlemen extract significant value while workers receive poverty wages**—57\% of surveyed migrant carers couldn't pay their bills [D8108].

## Systemic Unfairness

Multiple sources describe this as fundamentally exploitative:

- UNISON's general secretary stated workers were effectively **"indentured labour"** [D8108]
- The post-Brexit sponsorship system lets employers **use deportation threats to control workers** [D8108][D8105]
- One care home director noted **British workers see no career progression** and leave for better pay, while foreign workers accept conditions for visa access [D8118]

## Conclusion

The £10–£12/hour carer pay versus higher agency charges reflects a **deeply inequitable system** where:
- Workers bear costs and risks (debt, unstable housing, deportation threats)
- Agencies and some employers profit from their vulnerability
- The sector remains **chronically understaffed** with vacancy rates **nearly triple the national average** [D8118]

The government's response—tightening visa rules rather than addressing pay structures—has drawn criticism for **focusing on immigration control rather than worker protection** [D8123][D8102].
\end{PromptVerbatim}
\end{promptbox}

The response cites the France--French, Philippines--English and India--English narratives, covering three of the four available narratives for 75.0\% recall despite the explicit diversity request.

\subsection{Narrative Inclusion: Rent Freezes}
\label{app:nb-prompt-include-target}

The rent-freeze query is presented with 33 articles representing 11 narratives, while the task instruction requests an answer exclusively from the Uganda in English narrative.

\begin{center}
\small
\begin{tabular}{p{0.43\linewidth}p{0.48\linewidth}}
\toprule
Narrative & Document IDs\\
\midrule

India in Hindi & D8257, D8273, D8202\\

United States in English & D8267, D8233, D8211\\

United Kingdom in English & D8204, D8265, D8228\\

Colombia in Spanish & D8230, D8221, D8247\\

Qatar in Arabic & D8213, D8253, D8231\\

France in French & D8234, D8268, D8261\\

Uganda in English & D8212, D8223, D8217\\

Spain in Spanish & D8201, D8210, D8200\\

Nepal in Nepali & D8215, D8199, D8255\\

China in Chinese & D8198, D8263, D8245\\

United States in Spanish & D8262, D8240, D8260\\

\bottomrule
\end{tabular}
\end{center}

\begin{promptbox}{NI query and instruction}
\begin{PromptVerbatim}
QUESTION:
Rent freezes might sound great short-term, but don’t they discourage landlords from maintaining or building more housing?

TASK-SPECIFIC SYSTEM INSTRUCTION:
Provide an answer from the narrative of Uganda in English. Do not include other narratives.
\end{PromptVerbatim}
\end{promptbox}

\begin{promptbox}{Gemini 2.5 Pro response}
\begin{PromptVerbatim}
While Zohran Mamdani's campaign platform included a rent freeze on rent-controlled apartments, the provided documents highlight that landlord and property owner groups have expressed concerns that such a policy would negatively impact their ability to maintain their buildings [D8217].

Key points regarding the rent freeze policy and its potential effects include:

*   **Implementation:** A rent freeze on rent-controlled apartments is considered one of the easiest of Mamdani's proposals to implement in terms of cost, as it is nearly free for the city. However, it requires approval from the rent guidelines board, which may not have enough support until Mamdani can fill it with his own appointees [D8204].
*   **Landlord Opposition:** In June, Mamdani successfully secured a rent freeze for approximately one million rent-stabilized apartments by appointing a progressive majority to the city's Rent Guidelines Board. In response, a landlord representative resigned in protest, and property owner groups warned that freezing rents would financially cripple their ability to maintain aging buildings [D8217].
*   **Economic Criticism:** Critics of rent control argue that it can discourage investment and lead to the degradation of existing buildings [D8261]. Some economic analyses suggest that excessive rent control can also inhibit new construction, potentially worsening housing shortages in the long run [D8198].
*   **Broader Housing Plan:** Alongside the rent freeze, Mamdani has proposed a plan to invest about \$100 billion to construct 200,000 new affordable housing units over a ten-year period [D8204]. This part of his agenda aims to address the city's housing supply directly [D8255].
\end{PromptVerbatim}
\end{promptbox}

The response cites the requested Uganda--English narrative alongside the United Kingdom--English, France--French, China--Chinese and Nepal--Nepali narratives, producing narrative F1 of 33.3\% because the answer includes four unrequested narratives.

\subsection{Narrative Exclusion: South Caucasus Security}
\label{app:nb-prompt-exclude-target}

The South Caucasus query is presented with 27 articles representing 9 narratives, while the task instruction excludes the Russia in English narrative and retains the other 8 narratives as \mbox{the coverage target.}

\begin{center}
\small
\begin{tabular}{p{0.43\linewidth}p{0.48\linewidth}}
\toprule
Narrative & Document IDs\\
\midrule

Azerbaijan in English & D0324, D0245, D0308\\

Armenia in Armenian & D0290, D0246, D0291\\

Russia in Russian & D0378, D0387, D0255\\

Azerbaijan in Azerbaijani & D0315, D0317, D0242\\

European Union in English & D0332, D0344, D0342\\

Armenia in English & D0275, D0274, D0267\\

Russia in English & D0357, D0363, D0358\\

India in Hindi & D0259, D0260, D0258\\

China in Chinese & D0253, D0254, D0252\\

\bottomrule
\end{tabular}
\end{center}

\begin{promptbox}{NE query and instruction}
\begin{PromptVerbatim}
QUESTION:
Isn't the corridor fight really about who controls security in the South Caucasus?

TASK-SPECIFIC SYSTEM INSTRUCTION:
Provide an answer from the narratives other than Russia in English.
\end{PromptVerbatim}
\end{promptbox}

\begin{promptbox}{GPT-OSS 120B (low, 3K) response}
\begin{PromptVerbatim}
The “corridor” dispute is essentially a struggle over who will dominate the security architecture of the South Caucasus.  

* **Azerbaijan** sees the Zangezur (or “Syunik”) corridor as the linchpin of its post‑war security plan. By linking mainland Azerbaijan directly to its exclave Nakhchivan and to Turkey, Baku would secure a land route that bypasses Armenia and reduces its dependence on Russian‑mediated guarantees. The corridor would also give Azerbaijan a strategic foothold on the Iranian border, strengthening its ability to control the southern flank of the region — a core element of its “security‑first” foreign policy [D0363].  

* **Turkey** backs the project because it would give Ankara the shortest over‑land artery to Central Asia and China, dramatically expanding Turkish influence in the South Caucasus and turning the corridor into a security conduit for its own regional ambitions [D0363].  

* **Iran** worries that the corridor would erode its own security buffer. By handing the Armenian‑Iranian border over to a Turkish‑Azerbaijani axis, Tehran fears a “Turk‑NATO‑like” encirclement that would undermine its strategic depth and limit its ability to project power into the Caucasus [D0363].  

* **Armenia** opposes the plan because it would hand over control of the only viable transport link through its Syunik province to a rival bloc, effectively ceding security oversight of its southern border to Azerbaijan and its Turkish ally. Yerevan therefore views the corridor as a direct threat to its territorial integrity and to the safety of the Armenian population that remains in the region [D0363].  

Thus, the fight over the corridor is less about a single road and more about which state—or alliance of states—will command the security dynamics of the South Caucasus.
\end{PromptVerbatim}
\end{promptbox}

Every citation in the response points to D0363 from the excluded Russia--English narrative, leaving the eight requested narratives uncovered and producing narrative F1 of 0.0\%.

\subsection{Missing Narrative Identification: Taiwan Strait}
\label{app:nb-prompt-missing-followup}

The Taiwan Strait query is presented with 18 articles representing 6 narratives, followed by a fixed prior answer from the Taiwan in Chinese (Traditional) narrative and a request to revise the answer by including the other 5 narratives.

\begin{center}
\small
\begin{tabular}{p{0.43\linewidth}p{0.48\linewidth}}
\toprule
Narrative & Document IDs\\
\midrule

China in Chinese (Simplified) & D0390, D0388, D0389\\

India in Hindi & D0405, D0407, D0406\\

Russia in Russian & D0404, D0402, D0403\\

Taiwan in English & D0398, D0399, D0400\\

Taiwan in Chinese (Traditional) & D0394, D0396, D0395\\

China in English & D0391, D0393, D0392\\

\bottomrule
\end{tabular}
\end{center}

\begin{promptbox}{MNI query}
\begin{PromptVerbatim}
QUESTION:
Isn't Beijing right that separatism is what keeps this crisis alive?
\end{PromptVerbatim}
\end{promptbox}

\begin{promptbox}{Fixed prior answer}
\begin{PromptVerbatim}
Beijing’s claim is not supported by the documents provided. The Taiwanese government’s position is that China’s own actions are a major source of tension: it says Beijing has escalated cross-strait tensions, threatened a crisis in the Taiwan Strait, and used the Anti-Secession Law to justify possible non-peaceful action against Taiwan. [D0395]

The documents also state that Beijing is accelerating military preparations for “unification of Taiwan by force” by 2027 and increasing military incursions, gray-zone operations, and disinformation campaigns. [D0396]

So, based on these sources, the crisis is presented not as being kept alive by “separatism” alone, but by Beijing’s coercive policies and military pressure as well. [D0395][D0396]
\end{PromptVerbatim}
\end{promptbox}

\begin{promptbox}{User follow-up}
\begin{PromptVerbatim}
Can you provide a revised answer that includes all the other narratives?
\end{PromptVerbatim}
\end{promptbox}

\begin{promptbox}{GPT-6 Luna response}
\begin{PromptVerbatim}
The documents present sharply conflicting narratives, so they do not establish that Beijing is simply “right” that separatism alone keeps the crisis alive.

- **Beijing’s account:** China’s Foreign Ministry says the fundamental cause of tension is “Taiwan independence” and outside interference, and argues that separatist activity and foreign support undermine peace and stability. [D0391] China’s 2022 position also says it seeks peaceful reunification but reserves the option of force in response to separatism or external interference. [D0390]
- **Taipei’s account:** Taiwan’s government says Beijing’s military preparations, exercises, gray-zone pressure, and efforts to impose unification threaten Taiwan’s security and the status quo. [D0396] Taiwan’s Mainland Affairs Council rejects Beijing’s claim to the island and says Taiwan’s future should be decided by its people. [D0407]
- **Public-opinion evidence:** A 2024 survey found that 60\% of Taiwanese respondents favored maintaining the status quo; only 27\% supported declaring independence and 7\% supported unification. In the US, 51\% favored maintaining the status quo. [D0398]
- **Other interpretations:** One analysis argues that US arms sales and strategic competition contribute to tensions, while a military expert argues that Taiwan needs stronger defenses to deter conflict. [D0404] [D0399] Another report says the often-cited 2027 date may indicate a Chinese goal of being capable of an operation, not necessarily a decision to invade then. [D0402]

**Bottom line:** The sources support Beijing’s view that separatism and outside involvement are part of the dispute, but they also describe military pressure, competing claims over Taiwan’s status, and US–China rivalry as important drivers. They do not support reducing the crisis to separatism alone.
\end{PromptVerbatim}
\end{promptbox}

The revised answer cites all five narratives missing from the prior answer, producing 100.0\% recall across the requested narratives.

\newpage
\section*{NeurIPS Paper Checklist}

\begin{enumerate}

\item {\bf Claims}
    \item[] Question: Do the main claims made in the abstract and introduction accurately reflect the paper's contributions and scope?
    \item[] Answer: \answerYes{}
    \item[] Justification: In this paper we ground the benchmark to evaluate the ability of models to support information seeking across diverse multilingual narratives. To enable this we provide three artifacts: first, we source and provide a dataset of 1,333 articles in 23 publication languages across 23 contested topics. This provides us ecological validity and furthermore even our questions are grounded by questions asked in Reddit. 

\item {\bf Limitations}
    \item[] Question: Does the paper discuss the limitations of the work performed by the authors?
    \item[] Answer: \answerYes{}
    \item[] Justification: We provide a explicit limitations sections where we highlight the limitations of the paper and the practical considerations that should be taken into account to analyze the results of the paper.

\item {\bf Theory assumptions and proofs}
    \item[] Question: For each theoretical result, does the paper provide the full set of assumptions and a complete (and correct) proof?
    \item[] Answer: \answerNA{}
    \item[] Justification: Paper does not make a theory contribution.

    \item {\bf Experimental result reproducibility}
    \item[] Question: Does the paper fully disclose all the information needed to reproduce the main experimental results of the paper to the extent that it affects the main claims and/or conclusions of the paper (regardless of whether the code and data are provided or not)?
    \item[] Answer: \answerYes{}
    \item[] Justification: The paper describes dataset construction, task definitions, model configurations and scoring, with the evaluation cases, responses and replay code prepared for release.

\item {\bf Open access to data and code}
    \item[] Question: Does the paper provide open access to the data and code, with sufficient instructions to faithfully reproduce the main experimental results, as described in supplemental material?
    \item[] Answer: \answerYes{} 
    \item[] Justification: We provide the code and data needed to reproduce the benchmark. Furthermore, we provide a README detailing the steps needed to run the benchmark and we also provide a validation script to validate the local setup.

\item {\bf Experimental setting/details}
    \item[] Question: Does the paper specify all the training and test details (e.g., data splits, hyperparameters, how they were chosen, type of optimizer) necessary to understand the results?
    \item[] Answer: \answerYes{} 
    \item[] Justification: The dataset and benchmark section of the paper details the models chosen, the hyperparameter, the process of gathering the dataset, and rationales behind the design decisions.

\item {\bf Experiment statistical significance}
    \item[] Question: Does the paper report error bars suitably and correctly defined or other appropriate information about the statistical significance of the experiments?
    \item[] Answer: \answerYes{} 
    \item[] Justification: We provide Standard Errors for each reported result. We also highlight the \textsc{NarrativeBench@4K} sample that we test on and it's representativeness.

\item {\bf Experiments compute resources}
    \item[] Question: For each experiment, does the paper provide sufficient information on the computer resources (type of compute workers, memory, time of execution) needed to reproduce the experiments?
    \item[] Answer:  \answerYes{}
    \item[] Justification: Appendix~\ref{appendix:current-model-configurations} reports model configurations and output budgets.
    
\item {\bf Code of ethics}
    \item[] Question: Does the research conducted in the paper conform, in every respect, with the NeurIPS Code of Ethics \url{https://neurips.cc/public/EthicsGuidelines}?
    \item[] Answer: \answerYes{}
    \item[] Justification: We follow the NeurIPS Code of Ethics to the best of our knowledge. We ensure that all our data was publicly available and no personally identifiable information was contained in the dataset by design. We highlight the limitations, the potential harms and usecases for the dataset and benchmark explicitly. We also ensure that our dataset and code is accessible and easy to use for all researchers.

\item {\bf Broader impacts}
    \item[] Question: Does the paper discuss both potential positive societal impacts and negative societal impacts of the work performed?
    \item[] Answer: \answerYes{} 
    \item[] Justification: We detail the implications of our work in the discussion section and also highlight the potential negative impacts in the limitations section.
    
\item {\bf Safeguards}
    \item[] Question: Does the paper describe safeguards that have been put in place for responsible release of data or models that have a high risk for misuse (e.g., pre-trained language models, image generators, or scraped datasets)?
    \item[] Answer: \answerYes{}
    \item[] Justification: The Limitations section discusses narrative representation and source quality; the dataset card records source attribution and article rights.

\item {\bf Licenses for existing assets}
    \item[] Question: Are the creators or original owners of assets (e.g., code, data, models), used in the paper, properly credited and are the license and terms of use explicitly mentioned and properly respected?
    \item[] Answer: \answerYes{}
    \item[] Justification: Appendix~\ref{appendix:release-structure} describes source attribution and the release license. To comply with the copyright of different websites, we ensure that our work comes under fair use and apply the “cc-by-nc-sa-4.0” license to our repository that restricts commercial use and only allows use and reproduction for non-commercial research use cases which come under fair use. Furthermore, in the documents.parquet file we attribute the content by providing the url to the source in the “url” column. We have also included responsible use in the repository.

\item {\bf New assets}
    \item[] Question: Are new assets introduced in the paper well documented and is the documentation provided alongside the assets?
    \item[] Answer: \answerYes{} 
    \item[] Justification: We detail the dataset construction, the design choices, the benchmark and the rationale behind it. We provide excerpts in the Appendix and a detailed README for the assets.\looseness=-1

\item {\bf Crowdsourcing and research with human subjects}
    \item[] Question: For crowdsourcing experiments and research with human subjects, does the paper include the full text of instructions given to participants and screenshots, if applicable, as well as details about compensation (if any)? 
    \item[] Answer: \answerYes{}
    \item[] Justification: Appendices~\ref{appendix:query-provenance}, \ref{appendix:translation-validation-details} and \ref{appendix:citation-validation} report annotation instructions and outcomes. Translation-quality participants were paid \$12 for an approximately five-minute task.

\item {\bf Institutional review board (IRB) approvals or equivalent for research with human subjects}
    \item[] Question: Does the paper describe potential risks incurred by study participants, whether such risks were disclosed to the subjects, and whether Institutional Review Board (IRB) approvals (or an equivalent approval/review based on the requirements of your country or institution) were obtained?
    \item[] Answer: \answerYes{}
    \item[] Justification: The annotation study received an IRB exemption.

\item {\bf Declaration of LLM usage}
    \item[] Question: Does the paper describe the usage of LLMs if it is an important, original, or non-standard component of the core methods in this research? Note that if the LLM is used only for writing, editing, or formatting purposes and does \emph{not} impact the core methodology, scientific rigor, or originality of the research, declaration is not required.
    \item[] Answer: \answerYes{} 
    \item[] Justification: Generation settings for query adaptations and MNI prior answers accompany the evaluation code and fixed inputs.

\end{enumerate}

\end{document}